\documentclass{article}

\usepackage{PRIMEarxiv}

\usepackage[utf8]{inputenc} 
\usepackage[T1]{fontenc}    
\usepackage{hyperref}       
\hypersetup{hidelinks}

\usepackage{url}            
\usepackage{booktabs}       
\usepackage{amsfonts}       
\usepackage{nicefrac}       
\usepackage{microtype}      
\usepackage{lipsum}
\usepackage{fancyhdr}       
\usepackage{graphicx}       
\graphicspath{{media/}}     

\usepackage{calc}
\usepackage{indentfirst}
\usepackage{epstopdf}
\usepackage{lastpage}
\usepackage{ifthen}
\usepackage{lineno}
\usepackage{float}
\usepackage{amsmath}
\usepackage{setspace}
\usepackage{enumitem}
\usepackage{mathpazo}
\usepackage{booktabs} 
\usepackage[largestsep]{titlesec}
\usepackage{etoolbox} 
\usepackage{tabto} 
\usepackage[table]{xcolor} 
\usepackage{amsthm}

 \newcounter{problem}
\usepackage{authblk}
\title{Online learning of Riemannian hidden Markov models in homogeneous Hadamard spaces
}
\author[1]{\textbf{Quinten Tupker}}
\author[2]{\textbf{Salem Said}}
\author[3]{\textbf{Cyrus Mostajeran}}
\affil[1\;]{Centre for Mathematical Sciences, University of Cambridge, United Kingdom}
\affil[2\;]{CNRS, University of Bordeaux, France}
\affil[3\;]{Department of Engineering, University of Cambridge, United Kingdom}
\date{}                     
\begin{document}
\maketitle

\begin{abstract}
  Hidden Markov models with observations in a Euclidean space play an important role in signal and image processing. Previous work extending to models where observations lie in Riemannian manifolds based on the Baum-Welch algorithm suffered from high memory usage and slow speed. Here we present an algorithm that is online, more accurate, and offers dramatic improvements in speed and efficiency.
\end{abstract}


\section{Introduction}
\label{sec:Introduction}

Hidden Markov chains have played a major role in signal and image processing \cite{Cappe2006} with applications to image restoration \cite{Li2009markov}, speech recognition \cite{Rabiner1989} and protein sequencing \cite{Durbin1998}. The extensive and well-developed literature on hidden Markov chains is almost exclusively concerned with models involving observations in a Euclidean space. This paper builds on the recent work of \cite{said2021hidden} in developing statistical tools for extending current methods of studying hidden Markov models with observation in Euclidean space to models in which observations take place on Riemannian manifolds. Specifically, \cite{said2021hidden} introduces a general formulation of hidden Markov chain models with Riemannian manifold-valued observations and derives an expectation-maximization (EM) algorithm for estimating the parameters of such models. However, the use of manifolds often entails high memory usage. As such, here we investigate an ``online'' low memory alternative for fitting a hidden Markov process. Since this approach
only ever ``sees'' a limited subset of the data, it may be expected to inherently sacrifice some accuracy. Nonetheless, the dramatic increase in speed and efficiency often justifies such online algorithms.

The motivation for the development of hidden Markov models on manifolds and associated algorithms is derived from the wide variety of data encoded as points on various manifolds, which fits within the broader context of intense and growing interest in the processing of manifold-valued data in information science. In particular, covariance matrices are used as descriptors in an enormous range of applications including radar data processing \cite{Jeuris2016}, medical imaging \cite{Pennec2006}, and brain-computer interfaces (BCI) \cite{Barachant2012}. In the context of BCI, where the objective is to enable users to interact with computers via brain activity alone (e.g. to enable communication for severely paralysed users), the time-correlation of electroencephalogram (EEG) signals are encoded by symmetric positive definite (SPD) matrices \cite{Barachant2012}. Since SPD matrices do not form a Euclidean space, standard linear analysis techniques applied directly to such data are often inappropriate and result in poor performance. In response, various Riemannian geometries on SPD matrices have been proposed and used effectively in a variety of applications in computer vision, medical data analysis, and machine learning \cite{Log-Euclidean2006,Pennec2006}. In particular, the affine-invariant Riemannian metric has received considerable attention in recent years and applied successfully to problems such as EEG signal processing in BCI where it has been shown to be superior to classical techniques based on feature vector classification \cite{Barachant2012}. The affine-invariant Riemannian metric endows the space of SPD matrices of a given dimension with the structure of a Hadamard manifold, i.e. a Riemannian manifold that is complete, simply connected, and has non-positive sectional curvature everywhere \cite{lang2012}. In this work, we will restrict our attention to Hadamard manifolds, which encompass SPD manifolds as well as examples such as hyperbolic $d$-space.

\section{Hidden Markov chains}

In a hidden Markov model, one is interested in studying a hidden, time-varying,
finitely-valued process $(s_t; t = 1,2,\ldots)$ that takes values in a set $S$.
When $s_t=i$ for $i \in S$, we say that the process is in state
$i$ at time $t$. Moreover, we assume that the process is time-stationary,
so that there exists a transition matrix $A_{ij}$, which specifies the
conditional probabilities $ 
A_{ij}=\mathbb{P}(s_{t+1}=j|s_t=i)$.
If $\pi_i(t)=\mathbb{P}(s_t=i)$ is the distribution of $s_t$, then 
$\pi_{j}(t + 1)=\sum_{i\in S}\pi_i(t)A_{ij}$ describes the transition from time
$t$ to $t+1$. The states $s_t$ are ``hidden'', meaning that we can never know
their true values and can only observe them through random outputs $y_t$ that
take their values in a Riemannian manifold $M$ and are assumed to be generated
independently from each other. 

We assume that $M$ is a homogeneous Riemannian manifold that is also a Hadamard space, and that $y_t$ is distributed according to a Riemannian Gaussian distribution \cite{Salem2017} with probability density
\begin{equation}
p(y;c,\sigma)=\frac{1}{Z(\sigma)}\exp\left(-\frac{1}{2\sigma^2}d^2(y, c)\right),
\end{equation}
where $c\in M$ and $\sigma>0$ denote the mean and standard deviations of the distribution, respectively, and $Z(\sigma)$ is the normalization factor. In particular, on the space of $d\times d$ symmetric positive definite matrices $M=\mathcal{P}_d$, we have 
$\mathcal{P}_d=\mathrm{GL}(d)/\mathrm{O}(d)$,
where $\mathrm{GL}(d)$ is the general linear group of invertible $d\times d$ matrices, which acts transitively on $\mathcal{P}_d$ by $g\cdot y = gyg^{\dagger}$, where $g^{\dagger}$ denotes the transpose of $g$. The isotropy group is the space of $d\times d$ orthogonal matrices $\mathrm{O}(d)$. The affine-invariant Riemannian distance takes the form
\begin{equation} \label{Riemann distance}
d^2(y,z)=\operatorname{tr}\left[\left(\log(y^{-1}z)\right)^2\right],
\end{equation}
where $\operatorname{tr}$ denotes the trace operator and $\log$ the principal matrix logarithm. It is easy to see that this distance indeed satisfies the group invariance property $d(y,z) = d(g\cdot y, g\cdot z)$ for all $g\in G$, $y,z\in \mathcal{P}_d$.

Our assumption that $y_t$ is distributed according to a Riemannian Gaussian distribution is expressed as
\begin{equation}
p(y_t|s_t=i)=p(y_t;c_i,\sigma_i),
\end{equation}
where $p(y_t|s_t=i)$ denotes the conditional density with respect to the Riemannian volume measure. It follows that the probability density of $y_t$ is a time-varying mixture density
\begin{equation}
p(y_t)=\sum_{i\in S}\pi_i(t)p(y_t;c_i,\sigma_i).
\end{equation}
The objective is to estimate the transition matrix $(A_{ij})$ and the parameters $(c_i,\sigma_i)$ given access only to the observations $(y_t; t = 1,2,\ldots)$. This problem is addressed in \cite{said2021hidden} via an expectation-maximization (EM) algorithm using an extension of Levinson's forward-backward algorithm  \cite{Devijver1985,Rabiner1989}.

\section{Online estimation of Hidden Markov models}


The algorithm that we propose consists of a combination of an initialization phase, and a
fine-tuning phase. The initialization phase consists of running Riemannian
$K$-means on a limited subset of the data, while the fine-tuning phase is based
on an algorithm described by Krishnamurthy and Moore \cite{krishnamurthy1993}. Specifically,
 \cite{krishnamurthy1993} describes a hidden Markov model as a length $n$ chain $s_1, \dots s_n$ that switches
between $N$ different states, according to a transition matrix $A$, so
that $A_{ij} = \mathbb{P}(s_{k + 1} = j | s_k = i)$, and starts in an initial
state $\pi = \pi(1) \in \mathbb{R}^N$ given by $\pi_i = \mathbb{P}(s_1 = i)$. As before, we
assume that this Markov chain is ``hidden'', meaning that we never
know the true state $s_k$, but instead only see a representative $y_k$
of $s_k$. In our case specifically, we assume that $y_k$ is a
Riemannian Gaussian random variable $y_k \sim N(c_i, \sigma_i^2)$ with mean $c_i$ and standard
deviation $\sigma_i$. 

Initializing the algorithm using $K$-means on a limited subset of the data is
straightforward. After $K$-means has been completed using the Riemannian center of mass \cite{Bini2013}, we
count transitions between clusters to estimate the transition matrix, and
estimate the means $c_i$ of the Gaussian distributions as the means of the
clusters. Estimating the standard deviation of these Gaussian distributions is
more tricky. Here we introduce
\begin{equation}
  \delta = \frac{\partial}{\partial \eta} \log(Z(\eta)),
\end{equation}
where $\eta$ is the natural parameter $\eta = \frac{-1}{2\sigma^2}$. In the
Gaussian case, where $Z(\eta)$ is the normalization constant of the 
distribution, it is not hard to see that $\delta = \mathbb{E}(d^2(y, c))$, the
expected value of the square Riemannian distance from the Gaussian mean $c$ to
an observation $y$. In general, it can be quite challenging to compute $Z(\eta)$. Fortunately, recent work
by Santilli et al.\cite{santilli2020riemannian} outlines a method for calculating $Z(\eta)$ for SPD matrices in arbitrary dimension using orthogonal polynomials. In particular, \cite{santilli2020riemannian} provides explicit formulas for dimensions 2, 3, and 4, which could be used to establish a relationship between $\delta$ and $\sigma$, allowing us to estimate $\sigma$.
While this completes the initialization phase of the algorithm, we will continue to use this conversion frequently during the fine-tuning section of the algorithm as well.

For the fine-tuning step, we use a stochastic approximation method derived in \cite{krishnamurthy1993} based on the Kullback-Leibler information measure, which leads
to a stochastic gradient descent algorithm based on
\begin{equation}
  \lambda^{(k + 1)} = \lambda^{(k)} + \mathcal{J}^{-1}
  \partial_{\lambda^{(k)}} \log f(y_1, \dots, y_{k + 1} | \lambda^{(k)})
  \label{krishnamurthy update}
\end{equation}
where $\lambda^{(k)} = (A^{(k)}, c_i^{(k)}, \sigma_i^{(k)}, \pi_i^{(k)})$ is the $k$th estimate of the model parameters, $y_k$ is
the $k$th observation, $\mathcal{J}$ is the Fisher information matrix, and the
last derivative term can be called the score vector. 
Using superscript $k$ to denote the $k$th approximation of each quantity (so, for instance, $A^{(k)}$ is the $k$th approximation to the transition matrix), we collect definitions of the relevant conditional probabilities in Table \ref{krishnamurthy_equations}, where $f(x | y)$ denotes a conditional probability density function, $p_i(y)$ denotes the probability density function of the $i$th Gaussian
distribution, $P^{(k)}(s) = \operatorname{diag}(p_1^{(k)}(y_s), \dots, p_N^{(k)}(y_s))$, and $\mathbf{1}$ denotes the vector of ones.  Here $\Delta$ is the size of the ``minibatch'' of observations that the algorithm can see (so
stores in memory) at any
given moment, and the formulas given are the approximations used given the
limited size of the minibatch.

\begin{table}[ht]
\renewcommand{\arraystretch}{1.5}
\centering
\caption{\textit{Conditional probabilities}
\label{krishnamurthy_equations}}
\begin{tabular}{p{0.08\linewidth}>{\arraybackslash}p{0.33\linewidth}>{\centering\arraybackslash}p{0.51\linewidth}}
\toprule
Symbol & Definition & Calculation \\
\midrule\\
\addlinespace[-2ex]
   $\alpha_t(i)$ & $f(y_1, \dots, y_t, s_t = i | \lambda^{(t - 1)})$
                          & $ \alpha_t(j) = \sum_{i=1}^N \alpha_{t - 1}(i) A_{ij}^{(t - 1)} p_j^{(t - 1)}(y_t) ,$ \newline $  \alpha_1(i) = \pi_i(i) p_i^{(1)}(y_1)$  \\
                          \addlinespace[1.5ex]
      $\beta_{t|k}(i)$ & $f(y_{t + 1}, \dots, y_k | s_t = i, \lambda^{(k - 1)})$ &
      $\beta_{k|k+\Delta} = A^{(k - 1)} P^{(k - 1)}(k + 1) \dots A^{(k - 1)} P^{(k - 1)}(k
      + \Delta) \mathbf{1}$  \\ 
      \addlinespace[1.5ex]
      $\gamma_{t|k}(i)$ & $f(s_t = i | y_1, \dots, y_k,\lambda^{(k - 1)})$ &
      $\gamma_{t|k}(i) = \alpha_t(i) \beta_{t|k}(i) 
      /\sum_{j=1}^N\alpha_t(j) \beta_{t|k}(j)$ \\
      \addlinespace[1.5ex]
      $\zeta_{t|k}(i, j)$ & $f(s_t = i, s_{t+1} = j | y_1, \dots, y_k, \lambda^{(k - 1)})$ &
      $\frac{\alpha_t(i) A_{ij}^{(t - 1)} \beta_{t + 1|k}(j) p_j^{(t - 1)}(y_{t +
          1})}{\sum_i \sum_j \alpha_t(i) A_{ij}^{(t - 1)} \beta_{t + 1|k}(j) p_j^{(t - 1)}(y_{t + 1})}$ \\  
\bottomrule
\end{tabular}
\end{table}

To continue working out what Equation \eqref{krishnamurthy update} means in this
case, we find that 
for the Fisher information matrix of the transition matrix, it is helpful to define
\begin{equation}
  \mu_j^{(i)} = \frac{\sum_{t = 1}^{k + 1} \zeta_{t| k + 1} (i, j)}
  {(A^{(k + 1)}_{ij})^2},
\end{equation}
which is truncated to the size of the minibatch in practice. Similarly, for the
score vector of the transition matrix we find it useful to define
\begin{equation}
  g_j^{(i)} = \frac{\zeta_{k + 1| k + 1}(i, j)}{A_{ij}^{(k + 1)}},
\end{equation}
from which we can express our transition matrix update rule as
\begin{equation}
  A_{ij}^{(k + 1)} = A_{ij}^{(k)} + \frac{1}{\mu_j^{(i)}}
  \left( g_j^{(i)} -
    \frac{\sum_{h = 1}^N g_h^{(i)}/\mu_h^{(i)}}
        {\sum_{h = 1}^N 1 / \mu_h^{(i)}} \right).
\end{equation}
Similar work on the means and standard deviations yields the
following update rules when working with Gaussian distributions on the real line \cite{krishnamurthy1993}:
\begin{align}
  c_i^{(k + 1)} &= c_i^{(k)} + \frac{\gamma_{k + 1|k + 1}(i) (y_{k + 1} - c_i^{(k)})}
  {\sum_{t = 1}^{k + 1} \gamma_{t|k + 1}(i)} \\
  (\sigma_i^2)^{(k + 1)} &= (\sigma_i^2)^{(k)} + \frac{\gamma_{k + 1| k + 1}(i)
                    ((y_{k + 1} - c_i^{(k)})^2 - (\sigma_i^2)^{(k)})}{k + 1},
\end{align}
which, after adjusting the step sizes used, can
be converted to update rules on a Riemannian manifold as
\begin{equation}
  c_i^{(k + 1)} = c_i^{(k)} \#_\tau y_{k + 1}, \quad \quad \,\, \tau = \frac{\gamma_{k + 1|k + 1}(i)}
  {\sum_{t=1}^{k + 1} \gamma_{t|k + 1}(i)}, \\
\end{equation}
\begin{equation}
  \delta_i^{(k + 1)} = \delta_i^{(k)} + \frac{\gamma_{k + 1|k + 1}(i) \left(d^2\left(y_{k + 1}, c_i^{(k)}\right) -
    \delta_i^{(k)}\right)}{\sqrt{k}}, 
\end{equation}
where $x\#_\tau z$ denotes the unique point on the Riemannian geodesic from $x$ to $z$ that satisfies $d(x,x\#_\tau z)= \tau d(x,z)$ for $\tau\in[0,1]$. In particular, in the case of $\mathcal{P}_d$ equipped with the Riemannian distance given in Equation \eqref{Riemann distance}, we have
\begin{equation} \label{geodesics}
x\#_{\tau} z = x^{1/2}(x^{-1/2}zx^{-1/2})^{\tau}x^{1/2}.
\end{equation}

Finally, we may note that since these
intermediate calculations, particularly the conditional probabilities
mentioned, function as the ``memory'' of the algorithm, it is not sufficient to
initialize the algorithm by transferring only the values of $A, c_i, \delta_i$.
Instead, one must calculate values of $\alpha, \beta, \gamma, \zeta$ from the
$K$-means data as well. Fortunately, using the clusters obtained, one can do
this easily by using the final estimates obtained for $A, c_i, \delta_i$ from the
$K$-means algorithm.

\section{Computational experiment}
We consider a computational experiment comparing the EM
algorithm from \cite{said2021hidden} with our algorithm. For that, we generate a chain of length $10,000$ with values taken in a three-element set $S = \{1, 2, 3\}$, an initial distribution $\pi = \begin{pmatrix} 1 & 0 & 0 \end{pmatrix}$ (i.e.
certainly starting in state $i = 1$), and transition matrix
\begin{equation}
  A = (A_{ij}) =
  \begin{pmatrix}
    0.4 & 0.3 & 0.3 \\
    0.2 & 0.6 & 0.2 \\
    0.1 & 0.1 & 0.8
  \end{pmatrix}
\end{equation}
with means and standard deviations given by
\begin{align}
  c_1 &= 0 & c_2 &= 0.82i + 0.29 & c_3 &= 0.82 i - 0.29 \\
  \sigma_1 &= 0.2 & \sigma_2 &= 1 & \sigma_3 &= 1.
\end{align}
Here, the outputs $y_t$ are generated from a Riemannian Gaussian model in the
Poincar\'{e} disk model of hyperbolic $2$-space. That is, each $y_t$ takes values in $M = \{z \in
\mathbb{C} : |z| < 1 \}$, and
\begin{align} 
  d(y, z) &= \text{acosh} \left( 1 + \frac{2 |y - z|^2}{(1 - |y|^2)(1 - |z|^2)} \right) \label{Poincare distance} \\
  Z(\sigma) &= 2\pi\sqrt{\frac{\pi}{2}} \sigma e^{\frac{\sigma^2}{2}} \text{erf} \left( \frac{\sigma}{\sqrt{2}} \right) 
\end{align}
where $\operatorname{erf}$ denotes the error function. We use the Poincar\'{e} disk here rather
than $\mathcal{P}_2$ simply for ease of visualization. Moreover, the Poincar\'{e} disk with distance \eqref{Poincare distance} is isometric to the space of $2 \times 2$
symmetric positive definite matrices of unit determinant equipped with the affine-invariant Riemannian distance \eqref{Riemann distance}.


Applying our algorithm to this example, we obtain the results in Table \ref{accuracy_vs_Delta}.
We begin by comparing the speed and accuracy of our algorithm with the EM algorithm. Here, the online algorithm is
the clear winner, since it matches or exceeds the EM
algorithm on accuracy, and for $\Delta=200$ is around 450 times faster (all
tests performed on a standard personal laptop computer) --- a remarkable improvement. We observe a
pattern that accuracy decreases rapidly for very small $\Delta$ and is roughly
stable for $\Delta \geq 200$.
Also, if we only use $K$-means without any fine-tuning,  we see accuracy
is slightly lower (0.90), although it is quite fast (5s runtime). Finally, we
note that the runtimes scale sublinearly with $\Delta$, and in particular, the
scaling is very linear for small $\Delta$. Most likely it would be very linear
for all $\Delta$,
if it were not that the data that the $K$-means algorithm  (which is significantly
faster) uses is skipped in larger initialization batches, meaning that if $\Delta =
3,000$, then the (slower) fine-tuning algorithm runs only on the last $10,000 -
\Delta = 7,000$ data points, eventually leading to a sublinear curve.

\begin{table}[ht]
\renewcommand{\arraystretch}{1.5}
\centering
\caption{\textit{Online algorithm accuracy, runtime, and estimates for selected transition matrix elements for different minibatch sizes $\Delta$.}
\label{accuracy_vs_Delta}}
\begin{tabular}{p{0.16\linewidth}>{\centering\arraybackslash}p{0.1\linewidth}>{\centering\arraybackslash}p{0.15\linewidth}>{\centering\arraybackslash}p{0.1\linewidth}>{\centering\arraybackslash}p{0.1\linewidth}>{\centering\arraybackslash}p{0.1\linewidth}>{\centering\arraybackslash}p{0.1\linewidth}}
\toprule
Minibatch size, $\Delta$ & Accuracy & Runtime/s & 
$A_{11}$ & $A_{22}$  & $A_{33}$ & Transition RMSE  \\
\midrule\\
\addlinespace[-2ex]
True values &  &  & 0.4 & 0.6 & 0.8 & 0 \\
      40 & 0.48 & 1.37 & 0.40 & 0.30 & 0.40 & 1.49  \\ 
      60 & 0.50 & 1.93 & 0.31 & 0.34 & 0.28 & 1.39  \\ 
      80 & 0.76 & 2.54 & 0.36 & 0.49 & 0.45 & 1.26  \\ 
      100 & 0.86 & 3.21 & 0.48 & 0.63 & 0.59 & 1.13  \\ 
      200 & 0.98 & 5.81 & 0.46 & 0.60 & 0.77 & 1.12  \\ 
      300 & 0.94 & 8.39 & 0.57 & 0.63 & 0.75 & 0.95  \\ 
      1000 & 0.95 & 28.58 & 0.51 & 0.64 & 0.76 & 0.94  \\ 
      5000 & 0.95 & 70.69 & 0.41 & 0.58 & 0.70 & 0.91  \\ 
      $K$-means only & 0.90 & 4.99 & 0.53 & 0.65 & 0.56 & 0.93  \\ 
      EM  & 0.90 & 2623.69 & 0.31 & 0.88 & 0.96 & 1.29 \\ 
\bottomrule
\end{tabular}
\end{table}

Regarding the estimates of the transition matrix, we measure the error in the
transition matrix as the root mean squared error (RMSE) in the Frobenius norm
of the difference between the estimated and true transition matrices
$\lVert A - A' \rVert_{Frob}$. We consequently see that the EM algorithm underperforms all forms of the online algorithm for $\Delta < 200$. However, we also observe that the full
online algorithm rarely outperforms pure $K$-means significantly. In fact, when data
clusters are spread out as they are in this example, we find that the fine-tuning step does
not improve on $K$-means. However, for data clusters that do overlap
significantly, we have observed instances where the fine-tuning step improves
significantly on $K$-means. 

\begin{figure}
    \centering
    \includegraphics[width=\textwidth]{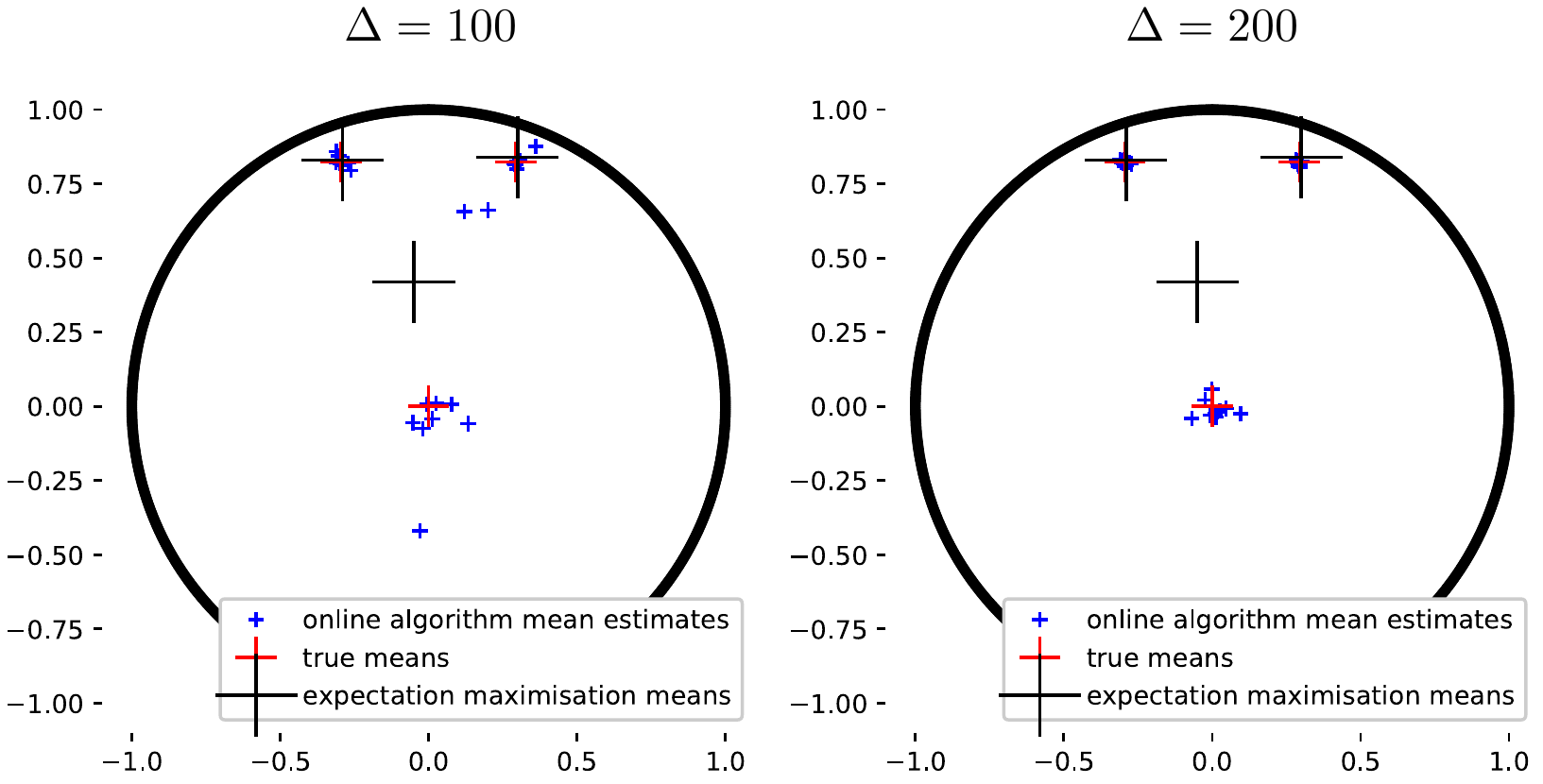}
    \caption{
    Mean estimates for multiple runs of the online algorithm for the example in the Poincar\'{e} disk for minibatch sizes of $\Delta=100$ and $\Delta = 200$ compared with the EM algorithm estimates. As expected, the quality of the online algorithm estimates improves with minibatch size $\Delta$. We observe that for a relatively modest minibatch size of $\Delta = 200$, the online algorithm is more accurate than the EM algorithm while being several orders of magnitude faster.
    }
    \label{poincare_by_Delta}
\end{figure}

Figure \ref{poincare_by_Delta} depicts results on mean estimation for minibatch sizes of $\Delta = 100$ and $\Delta = 200$. Note that the clustering of estimated means is more focused and closer to the true mean for $\Delta = 200$ than for $\Delta = 100$, as expected.
We also see that the EM mean appears to consistently
struggle to estimate the mean at the center of the disk. The degree of this error may be partially misconstrued since the Poincar\'{e} disk is significantly stretched near its edges compared to the center. Nevertheless, the EM algorithm certainly does seem to struggle here.

\section{Conclusion}


We have implemented an online algorithm for the learning of hidden Markov models on Hadamard spaces, i.e. an algorithm that uses a constant amount of memory instead of one that scales linearly with the amount of available data. We found that our algorithm outperforms a previous algorithm based on
expectation-maximisation in all measures of accuracy when the right minibatch size is used. Furthermore, this improvement is achieved while running nearly 450 times faster and using 50 times less memory in the example considered.

Future work will consider strategies that would allow the removal of the 
$K$-means initialization step and automatically tune the minibatch size without the need for experimentation.
Furthermore, significant challenges arise with increasing $N$, the number of states and hence dimension of the
transition matrix. Here difficulties include identifying the correct number of clusters, preventing the algorithm from merging separate clusters (undercounting), and
preventing the algorithm from double-counting clusters (overcounting) by assigning two means to
the same cluster. These challenges have proven to be the main obstacles in increasing $N$. By contrast, increasing the dimension of the Hadamard space in which observations take place (e.g. the space of positive definite matrices) does not pose significant challenges and is straightforward using recent work on computing normalization factors of Riemannian Gaussian distributions \cite{HSM21,santilli2020riemannian}. 

\section*{Appendix}

The update rules and equations that describe our algorithm are summarized in Table \ref{krishnamurthy_equations appendix}. Here $f(x | y)$ denotes a conditional probability density function,
$p_i(y)$ denotes the probability density function of the $i$th Gaussian
distribution, and $k$ denotes the iteration index of the algorithm, and any quantity labelled by $k$ such as $p_i^{k}(y)$ denotes the $k$th best estimate of $p_i(y)$. Finally, $\Delta$ is the size of the subset of data (the so-called ``minibatch'') that is loaded at
any given time. 

\begin{table}
\renewcommand{\arraystretch}{1.5}
\centering
\caption{\textit{Conditional probabilities and update rules}
\label{krishnamurthy_equations appendix}}
\begin{tabular}{p{0.084\linewidth}>{\arraybackslash}p{0.33\linewidth}>{\centering\arraybackslash}p{0.51\linewidth}}
\toprule
Symbol & Definition & Calculation \\
\midrule\\
\addlinespace[-2ex]
      $\alpha_t(i)$ & $f(y_1, \dots, y_t, s_t = i | \lambda^{(t - 1)})$
                          & $ \alpha_t(j) = \sum_{i=1}^N \alpha_{t - 1}(i) A_{ij}^{(t - 1)} p_j^{(t - 1)}(y_t) ,$ \newline $  \alpha_1(i) = \pi_i(i) p_i^{(1)}(y_1)$  \\
      \addlinespace[2ex]
      $\beta_{t|k}(i)$ & $f(y_{t + 1}, \dots, y_k | s_t = i, \lambda^{(k - 1)})$ &
      $\beta_{k|k+\Delta} = A^{(k - 1)} P^{(k - 1)}(k + 1) \dots A^{(k - 1)} P^{(k - 1)}(k
      + \Delta) \mathbf{1}$  \\ 
      \addlinespace[2ex]
      $\gamma_{t|k}(i)$ & $f(s_t = i | y_1, \dots, y_k,\lambda^{(k - 1)})$ &
      $\gamma_{t|k}(i) = \alpha_t(i) \beta_{t|k}(i) 
      /\sum_{j=1}^N\alpha_t(j) \beta_{t|k}(j)$ \\
      \addlinespace[2ex]
      $\zeta_{t|k}(i, j)$ & $f(s_t = i, s_{t+1} = j |$ \newline $y_1, \dots, y_k, \lambda^{(k - 1)})$ &
      $\frac{\alpha_t(i) A_{ij}^{(t - 1)} \beta_{t + 1|k}(j) p_j^{(t - 1)}(y_{t +
          1})}{\sum_i \sum_j \alpha_t(i) A_{ij}^{(t - 1)} \beta_{t + 1|k}(j) p_j^{(t - 1)}(y_{t + 1})}$ \\  
          \addlinespace[2ex]
      $\mu_j^i(k)$ & Intermediate calculation & $\sum_{t=1}^{k+1}
        \zeta_{t|k+1}(i, j)/(A^{(k+1)}_{ij})^{2}$   \\
        \addlinespace[2ex]
      $g_j^i(k + 1)$ & Intermediate calculation & $\zeta_{k + 1|k +
          1}(i, j)/A_{ij}^{(k+1)}$ \\ 
          \addlinespace[2ex]
      $c_i(k + 1)$ & Means updates &
      $c_i(k) \#_\tau y_{k + 1}, \qquad \tau = \gamma_{k + 1|k +1}(i)/\sum_{t=1}^{k + 1} \gamma_{t|k + 1}(i)$  \\  
            \addlinespace[2ex]
      $\delta_i^{(k + 1)}$ & $\delta$ update & $\delta_i^{(k)} + \frac{1}{\sqrt{k}}\gamma_{k + 1|k + 1}(i) \left(d^2\left(y_{k + 1}, c_i^{(k)}\right) -
    \delta_i^{(k)}\right)$ \\  
    \addlinespace[2ex]
      $A^{(k + 1)}$ & Transition matrix update & $A_{ij}^{(k)} + \frac{1}{\mu_j^{(i)}}
  \left( g_j^{(i)} -
    \frac{\sum_{h = 1}^N g_h^{(i)}/\mu_h^{(i)}}
        {\sum_{h = 1}^N 1 / \mu_h^{(i)}} \right)$ \\
\bottomrule
\end{tabular}
\end{table}

\section*{Acknowledgments}
This work benefited from partial support by the European Research Council under the Advanced ERC Grant Agreement Switchlet n.670645. Q.T. also received partial funding from the Cambridge Mathematics Placement (CMP) Programme. C.M. was supported by Fitzwilliam College and a Henslow Fellowship from the Cambridge Philosophical Society.

\bibliographystyle{splncs04}  
\bibliography{bibliography}

\end{document}